\pdfoutput=1

\documentclass[11pt]{article}

\usepackage{EMNLP2022}

\usepackage{times}
\usepackage{latexsym}

\usepackage[T1]{fontenc}

\usepackage[utf8]{inputenc}

\usepackage{microtype}

\usepackage{inconsolata}

\usepackage[utf8]{inputenc}

\usepackage{microtype}
\usepackage{rotating}
\usepackage{blindtext}
\usepackage{amsmath}
\usepackage{amsfonts}
\usepackage{graphicx}
\usepackage{mathtools}
\usepackage{booktabs}
\usepackage{relsize}
\usepackage{float}
\usepackage{graphics}
\usepackage{multirow}
\usepackage{algorithm}
\usepackage{algpseudocode}
\usepackage[utf8]{inputenc}

\usepackage{threeparttable}
\usepackage{booktabs, caption, makecell}

\usepackage[title]{appendix}
\usepackage[cal=cm]{mathalfa}
\usepackage[colorinlistoftodos]{todonotes}
\usepackage{multirow}
\usepackage[export]{adjustbox}
\usepackage{subcaption}
\usepackage{pifont}
\usepackage{tabularx}
\usepackage{longtable}
\usepackage{tikz}
\usepackage{enumitem}
\usepackage{diagbox}
\usepackage{bbm}

\usepackage{colortbl,dcolumn}
\usepackage{microtype}
\usepackage[subtle]{savetrees}
\title{SMaLL-100: Introducing \textbf{S}hallow Multilingual Machine Translation \textbf{M}odel for \textbf{L}ow-Resource \textbf{L}anguages}

\author{Alireza Mohammadshahi\thanks{~~~Work done during an internship at NAVER LABS Europe.}~~$^{1,2,3}$ ~~Vassilina Nikoulina$^1$ ~~Alexandre Berard$^1$ \\ \textbf{Caroline Brun}$^1$ ~~\textbf{James Henderson}$^2$ ~~\textbf{Laurent Besacier}$^1$ \vspace{0.1cm}\\ 
 $^1$ NAVER LABS Europe ~~~~~ $^2$ IDIAP Research Institute ~~~~~ $^3$ EPFL\\
 \texttt{\{first.last\}@naverlabs.com}\\\texttt{\{alireza.mohammadshahi,james.henderson\}@idiap.ch}
}

\begin{document}
\maketitle

\begin{abstract}

In recent years,  multilingual machine translation models have achieved promising performance on low-resource language pairs by sharing information between similar languages, thus enabling zero-shot translation.  To overcome the "curse of multilinguality", these models often opt for scaling up the number of parameters, which makes their use in resource-constrained environments challenging. We introduce \textit{SMaLL-100}, a distilled version of the M2M-100~(12B) model, a massively multilingual machine translation model covering 100 languages. We train SMaLL-100 with uniform sampling across all language pairs and therefore focus on preserving the performance of low-resource languages.  We evaluate SMaLL-100 on different low-resource benchmarks: FLORES-101, Tatoeba, and TICO-19 and demonstrate that it outperforms previous massively multilingual models of comparable sizes (200-600M) while improving inference latency and memory usage. Additionally, our model achieves comparable results to M2M-100~(1.2B), while being 3.6$\times$ smaller and 4.3$\times$ faster at inference.\footnote{The code and pre-trained SMaLL-100 model is available at \url{https://github.com/alirezamshi/small100}.}

\end{abstract}

\section{Introduction}

Neural Machine Translation~(NMT) systems are usually trained on datasets consisting of millions of parallel sentences, thus still performing poorly on low-resource languages, i.e., languages without a large amount of training data. 
Over the past few years, previous work has proposed several approaches to improve the quality of translations in low-resource languages, e.g., Multilingual Neural Machine Translation~(MNMT) models~\cite{johnson-etal-2017-googles,m2m-100,tang-etal-2021-multilingual,flores101}, back-translation~\cite{sennrich-etal-2016-improving,edunov-etal-2018-understanding} and unsupervised machine translation~\cite{garcia-etal-2021-harnessing,ko-etal-2021-adapting}. 
Massively MNMT models are particularly interesting for low-resource languages as they benefit the most from knowledge transfer from related languages~\cite{https://doi.org/10.48550/arxiv.1907.05019}. However, it is also seen that \textit{curse of multilinguality} hurts the performance of high-resource languages. So, previous work attempted to increase the model size to maintain the translation performance in both high and low-resource languages. This makes the use of these massively MNMT models challenging in real-world resource-constrained environments. To overcome this problem, we propose SMaLL-100, a \textbf{S}hallow \textbf{M}ultilingual M\textbf{a}chine Translation Model for \textbf{L}ow-Resource \textbf{L}anguages covering 100 languages, which is a distilled alternative of M2M-100~(12B)~\cite{m2m-100}, the most recent and biggest available multilingual NMT model. In this paper, we focus on very-low and low-resource language pairs as there is no reasonable-size universal model that achieves acceptable performance over a great number of low-resource languages. We do so by training SMaLL-100 on a perfectly balanced dataset.\footnote{All language pairs have the same sampling probability, regardless of their training data size.} While this leads to lower performance on the high-resource languages, we claim that this loss is easily recoverable through further fine-tuning.  We evaluate SMaLL-100 on different low-resource benchmarks, e.g., FLORES-101~\cite{flores101}, Tatoeba~\cite{tiedemann-2020-tatoeba}, and TICO-19~\cite{anastasopoulos-etal-2020-tico}. To summarize, our contributions are as follows:
\begin{itemize}[noitemsep,topsep=0pt,parsep=0pt,partopsep=0pt]
\item We propose SMaLL-100, a shallow multilingual NMT model, focusing on low-resource language pairs.
\item We evaluate SMaLL-100 on several low-resource NMT benchmarks.
\item We show that our model significantly outperforms previous multilingual models of comparable size while being faster at inference. Additionally, it achieves comparable results with M2M-100~(1.2B) model, with 4.3$\times$ faster inference and a 3.6$\times$ smaller size.
\item While SMaLL-100 reaches 87.2\% performance of the 12B teacher model, we show that this gap can be closed with a few fine-tuning steps both for low and high-resource languages. 
\end{itemize}

\section{Model and Training}
\label{sec:method}
\newcommand{\Lagr}{\mathcal{L}}

\subsection{SMaLL-100 Architecture} 
It has been shown by \citet{kasai2021_deepshallow} that deep encoder / shallow decoder architectures can achieve good translation quality while being significantly faster at inference. \citet{berard-etal-2021-efficient} have confirmed that this is also valid for multilingual NMT. Here, we use a 12-layer Transformer encoder \cite{transformer} and 3-layer decoder. Table \ref{app:tab:hyper-params} in the Appendix \ref{app:impl} reports further details of the SMaLL-100 architecture. Different from M2M-100 model, we use language codes in the encoder side, as it is shown to perform better with shallow decoder architectures~\cite{berard-etal-2021-efficient}.  

\subsection{Training Strategy}
SMaLL-100 is trained with a combination of two loss functions: a standard Cross Entropy loss (CE) and a Knowledge Distillation loss (KD). Given source sequence $X$ and gold target translation $Y=(y_0,...,y_m)$, the CE loss is calculated as:
\begin{align}
\begin{split}
\Lagr_{ce} = - \sum_{j=0}^{m}\sum_{z=1}^{|K|} \mathbbm{1}\{y_j = z \} \operatorname{log}~p(y_j=z | y_{<j},X,\theta_S) 
\end{split}
\end{align}
where $|K|$ is the target vocabulary size, $\mathbbm{1}$ is the indicator function, and $\theta_S$ is the model parameters. $p()$ is the conditional probability function. \\
We additionally use a word-level distillation loss, which is the Kullback–Leibler divergence between the output distributions of the student and teacher models~\cite{hu-etal-2018-attention}. Specifically, it is calculated as:
\begin{align}
\begin{split}
\Lagr_{kd} = - \sum_{j=0}^{m}\sum_{z=1}^{|K|} q(y_j=z | y_{<j},X,\theta_T) \\
\times \operatorname{log}~p(y_j=z | y_{<j},X,\theta_S) 
\end{split}
\end{align}
where $\theta_T$ is parameters of the teacher model. $q()$ is the conditional probability of the teacher model. The total loss is computed as: 
\begin{align}
\begin{split}
\Lagr_{total} = \Lagr_{ce} + \alpha \Lagr_{kd}
\end{split}
\end{align}
where $\alpha$ is a trainable parameter. 

\subsection{Training Data} 
Our training data includes parallel sentences from CCMatrix~\cite{https://doi.org/10.48550/arxiv.1911.04944} and CCAligned~\cite{el-kishky-etal-2020-ccaligned} datasets, which are part of the training data used by \citet{m2m-100} to train the M2M-100 models. As our goal is to maintain the performance of low-resource languages, we balance the training data across all language pairs; specifically, 100K sentence pairs are sampled for each language pair.\footnote{For language pairs with less than 100K sentence pairs, we repeat their data. We randomly select 100K sentences for language pairs with more than 100K training sentences.} As a result, our training data contains nearly 456M parallel sentences, which is less than 6$\%$ of the original data on which M2M-100~\cite{m2m-100} was trained. We use the same languages as M2M-100.
\section{Experimental Setup}
\label{sec:setup}

\begin{table}
\centering
\begin{adjustbox}{width=0.7\linewidth}
\begin{tabular}{lc}

Resource Type & Criteria \\
\toprule
Very-Low & $|K| \leq 100K$  \\
Low & $100K < |K| \leq 1M $ \\
Medium & $1M < |K| \leq 100M $ \\
High & $100M < |K|$ \\
\bottomrule
\end{tabular}
\end{adjustbox}
\caption{\label{tab:langdist} The criteria to split languages into different resource categories. $|K|$ is the amount of training data to/from English.}
\end{table}

\begin{table*}
	\begin{adjustbox}{width=\textwidth}
	\centering
	\begin{tabular}{lcc|cccccccccccc|c}\toprule
		 &\multicolumn{15}{c}{\hspace{2cm} Language Direction} \\
		\cmidrule{4-15} 
		Model & params & Speed & VL2VL & VL2L & VL2M & VL2H & L2VL & L2L & L2M & L2H & M2VL & M2L & H2VL & H2L & AVG \\ \midrule
		\textbf{FLORES-101} & & & &  &  &  & &  &  &  &  & &  & &  \\[0.1ex] 
		FLORES-124 & 175M & 5.3$\times$ & 3.3 & 3.4 & 6.0 & 7.8 & 3.7 & 3.1 & 6.9 & 8.8 & 6.9 & 5.2 & 8.1 & 6.0 & 5.8\\
		M2M-100 & 418M & 3.1$\times$ & 4.3 & 3.7 & 7.8 & 9.4 & 5.4 & 3.4 & 9.1 & 11.3 & 9.9 & 5.8 & 11.4 & 6.6 & 7.3 \\
		FLORES-124 & 615M & 2.9$\times$ & 5.1 & 5.1 & 9.2 & 11.2 & 5.8 & 4.7 & 10.6 & 13.1 & 10.3 & 7.6 & 11.5 & 8.5 & 8.6 \\
	    \textit{Finetuned-100} & 330M & 7.8$\times$ & 6.1 & 5.4 & 8.7 & 11.3 & 5.7 & 4.1 & 9.0 & 11.8 & 10.4 & 6.8 & 13.0 & 8.0 & 8.4 \\
	    \textit{SMaLL-100} & 330M & 7.8$\times$ & \underline{7.9} & \underline{7.0} & 10.3 & 12.6 & 8.4 & \underline{6.1} & 11.6 & 14.3 & \underline{13.7} & \underline{9.0} & \underline{16.7} & \underline{10.2} & \underline{10.7} \\
	    \midrule
	    M2M-100 & 1.2B & 1.8$\times$ & 6.7 & 6.1 & \underline{10.8} & \underline{12.8} & \underline{8.7} & \underline{6.1} & \underline{13.0} & \underline{15.9} & 13.6 & 8.8 & 15.4 & 9.7 & 10.6 \\
	    M2M-100 & 12B & 1$\times$ & \textbf{8.7} & \textbf{8.8} & \textbf{11.9} & \textbf{13.7} & \textbf{11.7} & \textbf{9.7} & \textbf{15.4} & \textbf{18.2} & \textbf{16.5} & \textbf{12.6} & \textbf{18.7} & \textbf{13.9} & \textbf{13.3} \\
          \midrule \midrule
	   \textbf{Tatoeba} & & & &  &  &  & &  &  &  &  & &  & &  \\[0.1ex] 
		FLORES-124 & 175M  & 5.3$\times$ & - &7.6 & 15.7 & 10.1 & 4.6 & 5.3 & 11.5 & 10.8 & 14.0 & 10.2 & 6.4 & 7.5 & 9.4 \\
		M2M-100 & 418M  & 3.1$\times$ &- & 7.4 & 19.7 & 12.3 & 5.9 & 5.3 & 13.8 & 13.2 & 14.9 & 11.7  & 7.7 & 9.0 & 10.9 \\
		FLORES-124 & 615M  & 2.9$\times$ & - & \textbf{9.1} & 19.4 & 11.4 & 6.9 & \underline{7.6} & 12.7 & 13.7 & 14.4 & 13.3 & 8.0 & 9.7  & 11.4 \\
	    \textit{Finetuned-100} & 330M  & 7.8$\times$ & - & 4.0 & 21.1 & \underline{14.4} & 7.7 & 5.2 & 15.3 & 14.2 & 14.0 & 12.1 & 8.9 & 8.3 & 11.4 \\
	    \textit{SMaLL-100} & 330M  & 7.8$\times$ & - & 4.6 & \underline{22.1} & \textbf{16.4} & \underline{8.7} & 7.0 & \underline{16.7} & 15.8 & 16.3 & \underline{14.5} & \underline{10.6} & \underline{11.2} & \underline{13.1} \\
	    \midrule
	    M2M-100 & 1.2B  & 1.8$\times$ &- & \underline{8.8} & 19.5 & 13.1 & \underline{8.7} & 7.2 & 16.3 & \underline{17.0} & \underline{17.2} & 13.4 & \textbf{10.7} & 11.1  & 13.0 \\
	    M2M-100 & 12B & 1$\times$ & - & 8.6 & \textbf{23.5} & 13.1 & \textbf{9.8} & \textbf{10.2} & \textbf{17.8} & \textbf{17.9} & \textbf{18.5} & \textbf{15.2} & \textbf{10.7} & \textbf{13.2} & \textbf{14.4} \\ \midrule \midrule

	    \textbf{TICO-19} & & & &  &  &  & &  &  &  &  & &  & &  \\[0.1ex] 
		FLORES-124 & 175M & 5.3$\times$ & 4.6 & 5.5 & 8.1 & 11.5 & 4.4 & 5.6 & 9.7& 12.2& 3.9& 8.0& 4.2& 8.7 & 7.2\\
		M2M-100 & 418M & 3.1$\times$ & 4.0 & 5.5 & 9.8 & 13.7 & 4.2 & 5.7 & 11.6 & 14.9 & 4.1 & 8.8 & 5.3 & 9.4 & 8.1 \\
		FLORES-124 & 615M & 2.9$\times$ &  4.6 & 7.4 & 11.5 & 16.4 & 4.8 & 7.6 & 12.9 & 16.7 & 4.4 & 10.7 & 4.4 & 11.5 & 9.4 \\
	    \textit{Finetuned-100} & 330M & 7.8$\times$ & 6.1 & 7.2 & 11.9 & 17.4 & 5.5 & 6.1 & 12.1 & 15.2 & \underline{6.4} & 9.0 & \underline{9.5} & 10.3 & 9.7 \\
	    \textit{SMaLL-100} & 330M & 7.8$\times$& \textbf{7.8} & \underline{8.8} & \underline{13.3} & \underline{19.0} & \textbf{8.0} & 8.5 & \underline{14.3} & 17.8 & \textbf{8.3} & \underline{11.5} & \textbf{11.3} & \underline{12.7} & \underline{11.8} \\ 
	    \midrule
	    M2M-100 & 1.2B & 1.8$\times$ & 5.4 & 8.2 & 13.2 & 18.9 & 6.0 & \underline{8.7} & 14.0 & \underline{19.2} & 5.2& \underline{11.5} & 6.1& 12.5 & 10.8 \\
	    M2M-100 & 12B & 1$\times$ & \underline{6.4} & \textbf{10.9} & \textbf{15.4} & \textbf{20.6} & \underline{7.8} & \textbf{11.9} & \textbf{16.6} & \textbf{21.4} & \underline{6.4} & \textbf{15.4} & 8.7& \textbf{16.4} & \textbf{13.1} \\
	    \midrule
		\bottomrule
	\end{tabular}
	\end{adjustbox} 
	\caption{\label{table:main} Average spBLEU performance on FLORES-101, Tatoeba, and TICO-19 benchmarks for different language pair categories, defined in Appendix~\ref{app:detail-eval}. FLORES-101 results are computed on language pairs where M2M-100 12B has spBLEU scores higher than 3 to avoid polluting the analysis with meaningless scores. The first and second columns give the model size and speed-up ratios compared to M2M-100~(12B). Last column is the average spBLEU performance over all mentioned language directions. The best scores are shown in bold, and the second best results are shown with underline.}
\end{table*}

\subsection{Evaluation Benchmarks}

\paragraph{FLORES-101} is a multilingual NMT benchmark, containing 3,001 sentences from different domains, that are derived from English Wikipedia. Sentences are translated into 101 languages by human translators~\cite{flores101}. It mostly includes low and medium-resource languages. We use {\tt devtest} subset for the evaluation.

\paragraph{Tatoeba} is a crowd-sourced collection of user-provided translations in different languages~\cite{tiedemann-2020-tatoeba}. We choose a subset of languages from {\tt test} set of Tatoeba Challenge,\footnote{\url{https://github.com/Helsinki-NLP/Tatoeba-Challenge}} which are covered by M2M-100. 

\paragraph{TICO-19} was created during the COVID-19 pandemic~\cite{anastasopoulos-etal-2020-tico}. It contains sentences from 36 languages in the medical domain, including 26 low-resource languages. We evaluate on languages which are covered by M2M-100~\cite{m2m-100}. 

Inspired by \citet{flores101}, we split the languages based on the amount of available training sentences aligned with English into 4 different categories: Very-Low~(VL), Low~(L), Medium~(M), and High-resource~(H). As the true amount of training data is both dependent on quality and quantity of parallel sentences, \citet{flores101} suggested to estimate it by computing the number of bitext data aligned with English, that is calculated from statistics of OPUS corpora~\cite{tiedemann-2012-parallel}. Table~\ref{tab:langdist} illustrates the criteria for choosing the category of different languages. More details about the distribution of language pair categories in each benchmark are provided in Appendix~\ref{app:detail-eval}. 

\subsection{Baselines}

\paragraph{M2M-100}\citet{m2m-100} is a recent many-to-many NMT model covering 100 languages. \citet{m2m-100} provide 3 variants with respectively 418M, 1.2B, and 12B parameters. We compare against these 3 variants.
\paragraph{FLORES-124} is an extension of M2M-100, covering additional 24 languages. Training data of the additional languages is derived from OPUS~\cite{tiedemann-2012-parallel}. \citet{flores101} provide two models with 175M and 615M parameters. We use both models as baselines. 
\paragraph{FineTuned-100} uses the same architecture as defined in Section~\ref{sec:method}, but KD loss~($\Lagr_{kd}$) is not used for training. For a fair comparison, it is trained for the same number of steps as SMaLL-100 model. 

\begin{table*}
\centering
\begin{adjustbox}{width=0.9\linewidth}
\begin{tabular}{lccccc}\toprule
    & &\multicolumn{3}{c}{Fine-tuned} \\
    \cmidrule{3-5} 
Language pair & Language Type & M2M-100 (418M) & SMaLL-100 & steps & M2M-100 (12B) \\ \midrule
Cebuano-English & Low & 18.6 & \textbf{29.8} & 0.5K & 27.7   \\
English-Igbo & Low & 9.2 & \textbf{15.7}  & 1.5K & 14.9 \\
English-Malayalam & Low & 16.7 & \textbf{20.8}  & 3K & 20.6 \\
Georgian-Russian & Medium & 6.9 & \textbf{13.1} & 0.5K & 10.1 \\
English-Italian & High & 33.3 & 33.4 & 20K & \textbf{33.5} \\
French-Italian & High & 31.3 & 31.5 & 20K & \textbf{32.0} \\
Italian-Spanish & High & 26.6 & 26.8 & 20K & \textbf{27.0} \\
\bottomrule
\end{tabular}
\end{adjustbox}
\caption{\label{app:fine-tune} spBLEU performance of fine-tuned SMaLL-100 and M2M-100 (418M) for the specified step, and M2M-100~(12B) on FLORES-101 devtest. The "step" column is the number of training steps required to reach M2M-100~(12B) performance. The type of each language pair is defined as the minimum of source and target language categories.}
\end{table*}

\subsection{Implementation Details}

SMaLL-100 contains nearly 330M parameters with 12 encoder and 3 decoder Transformer layers.\footnote{It is initialized with M2M-100~(418M), using its first 3 decoder layers for the initialization of the student's decoder.} It is trained for 30 days on 16 TESLA V100-32GB GPUs,\footnote{4 GPUs are used for the training of the student model. 12 GPUs are utilized to do the model parallelism of the teacher model.} with a batch size of 1K tokens and accumulated gradients over 9 batches. We implement our model using fairseq repository.\footnote{\url{https://github.com/facebookresearch/fairseq}} We use {\tt last-checkpoint}\footnote{\url{https://github.com/facebookresearch/fairseq/tree/main/examples/m2m_100}} of M2M-100~(12B) for the teacher model. For decoding, the beam search of 5 is applied. All hyper-parameters regarding the architecture and optimization strategy are provided in Appendix~\ref{app:impl}. \\ 
For a faster convergence, we first fine-tune SMaLL-100 for 150k steps without distillation~($\Lagr_{kd}$). Then, it is trained with both losses for 756K steps~(nearly 1 epoch). For evaluation, we use SentencePiece BLEU~(spBLEU), as it is shown to be a fair metric in multilingual settings~\cite{flores101}.\footnote{It utilizes a SentencePiece tokenizer with 256K tokens:~\url{https://github.com/facebookresearch/flores}} We use the same tokenizer and dictionary as M2M-100.

\section{Results and Discussion}
\label{sec:result}

\subsection{Low-Resource NMT Benchmarks}

Table~\ref{table:main} shows the average spBLEU performance on FLORES-101, Tatoeba, and TICO-19 test sets for different categories of language directions.\footnote{Complete spBLEU calculations of different language pairs on tested NMT benchmarks are provided in Appendix~\ref{app:spbleu}. Speed is calculated on 2 TESLA V100-32GB GPUs with a batch size of 1 sentence over a subset of FLORES-101 devtest, containing nearly 10K sentences from all language pairs.} SMaLL-100 outperforms all the models with comparable sizes while being smaller and faster at inference. Specifically, it outperforms M2M-100 418M both in terms of performance (+3.1 spBLEU) and inference speed (2.5$\times$ faster). We believe that Finetuned-100 outperforms M2M-100 418M for low-resource languages thanks to finetuning on the balanced dataset.  The higher performance of SMaLL-100 compared to Finetuned-100 across all benchmarks shows the benefit of KD loss which allows to distill knowledge from the teacher model.  Additionally, SMaLL-100 achieves competitive results with M2M-100~(1.2B), while being 3.6$\times$ smaller and 4.3$\times$ faster at inference. Compared to the biggest M2M-100 model~(12B), SMaLL-100 loses nearly 1.7 spBLEU but is 36$\times$ smaller and 7.8$\times$ faster. Regarding medium and high-resource language pairs~(as shown in Appendix~\ref{app:spbleu:remain}), SMaLL-100 achieves better or similar performance compared to M2M-100~(418M) and FLORES-124~(615M), while it contains fewer parameters and is faster at the evaluation time. It under-performs for some medium and high-resource language pairs compared to the teacher model~(M2M-100 12B), which could be easily recovered, as we describe in the remaining section.

\subsection{Recovering Teacher Model Performance}
To go further, we demonstrate that SMaLL-100 can easily recover the performance of the teacher model with just a few fine-tuning steps, both for low and high-resource language pairs. For comparison, we fine-tune M2M-100 (418M) model with the same number of steps. \\
Table~\ref{app:fine-tune} reports spBLEU performance for several language pairs, alongside the number of fine-tuning steps, required by SMaLL-100 model to reach M2M-100 (12B) performance.\footnote{More dataset and implementation details of these fine-tuning experiments are provided in Appendix~\ref{app:detail-low}.}
We see that SMaLL-100 achieves better performance than M2M-100~(12B) after a few training steps on low-resource language pairs. For high-resource language pairs, SMaLL-100 is fine-tuned for 20K steps to reach the performance of M2M-100~(12B) model. Additionally, fine-tuned SMaLL-100 significantly outperforms fine-tuned M2M-100 (418M) model on low and medium-resource languages. This confirms that SMaLL-100 could be a powerful and lightweight initialization model for training on different language pairs.

\section{Related Work}
\label{relatedwork}

\paragraph{Compression and Distillation.} Over the past few years, pre-trained models lead to significant improvement by increasing the parameter size~\cite{raffel2019exploring,m2m-100,zhang2022opt}, which makes it challenging to use them in the resource-constraint environment. Previous work use several compression techniques e.g. knowledge distillation~\cite{kim-rush-2016-sequence,light2021}, pruning~\cite{behnke-heafield-2020-losing,zhang-etal-2021-enlivening,mohammadshahi2022compressed}, and quantization~\cite{tao2022compression,yao2022zeroquant} to provide a reasonable-size model, while keeping the performance. 
\paragraph{Multilingual NMT.} It provides a single model to translate between any pair of languages, which significantly improves performance on low-resource languages thanks to knowledge transfer~\cite{haddow2021survey}. 
Several works~\cite{dong-etal-2015-multi,firat-etal-2016-multi,platanios-etal-2018-contextual,m2m-100,berard-etal-2021-efficient} propose to include both language-specific, and language-independent parameters in MNMT models. Recently, massively MNMT models~\cite{neubig-hu-2018-rapid,https://doi.org/10.48550/arxiv.1907.05019,aharoni-etal-2019-massively,m2m-100,zhang-etal-2020-improving} have been proposed to translate between more than 100 languages. However, these models usually contain a huge number of parameters to maintain performance in both high and low-resource languages. Different from the previous work, we introduce SMaLL-100, which outperforms previous models with comparable size in low-resource language directions, while achieving better speed and being smaller.

\section{Conclusion}

We presented SMaLL-100 model, a shallow multilingual NMT model, focusing on low-resource languages. We evaluated our model on different NMT benchmarks.
SMaLL-100 significantly outperforms multilingual models of comparable size on all of the tested benchmarks (FLORES-101, Tatoeba, TICO-19) and is much faster at inference.
It also achieves competitive results with M2M-100 1.2B~\cite{m2m-100}, while being 4.3$\times$ faster at inference and 3.6$\times$ smaller. Compared to M2M-100~(12B), the biggest available MNMT model, SMaLL-100 loses nearly 1.7 spBLEU on average but it is significantly faster~(7.8$\times$) and smaller~(36$\times$), which makes it a good fit for resource-constrained settings. Additionally, we show that SMaLL-100 can achieve similar performance as M2M-100~(12B) with just a few steps of fine-tuning on specific language pairs.
\section*{Limitations}

As mentioned in Section~\ref{sec:result}, SMaLL-100 model under-performs for some medium and high-resource languages, which could be resolved by further fine-tuning. Due to our computation constraint, we train SMaLL-100 model on nearly 6$\%$ of the original M2M-100 model. So, we encourage future research to increase the size of training data~(especially for low-resource languages) to achieve better performance. Also, future research could apply different distillation strategies~\cite{wu-etal-2020-skip,wang-etal-2021-selective}, as we just used word-level knowledge distillation loss~\cite{hu-etal-2018-attention}.
\section*{Acknowledgement}

The work is done during the research internship at NAVER LABS Europe. Alireza Mohammadshahi is supported by the Swiss National Science Foundation~(grant number CRSII5-180320).

\bibliographystyle{acl_natbib}
\bibliography{emnlp2021}

\newpage
\renewcommand\thesection{\Alph{section}}
\renewcommand\thesubsection{\thesection.\Alph{subsection}}
\setcounter{section}{0}
\onecolumn
\begin{appendices}

\section{Details of Evaluation benchmarks}
\label{app:detail-eval}

\subsection{Resource-type of Languages in Evaluated Datasets}
\label{app:lang-type}

\begin{table}[!ht]
\centering
\begin{adjustbox}{width=0.7\linewidth}
\begin{tabular}{cc|cc|cc|cc|cc}\toprule
af & Low & lg & Very-Low & lt & Medium & sn & Low & gl & Medium \\ \midrule
am & Low & ka & Medium & luo & Low & sd & Very-Low & gd & Low \\ \midrule
ar & Medium & de & High & lb & Medium & sk & Medium & ht & Low \\ \midrule
hy & Low & el & Medium & mk & Medium & sl & Medium & su & Low \\ \midrule
as & Very-Low & gu & Low & ms & Low & so & Low & ln & Very-Low \\ \midrule
ast & Low & ha & Low & ml & Low & ku & Low & ilo & Low \\ \midrule
az & Low & he & Medium & mt & Medium & es & High & mg & Medium \\ \midrule
be & Very-Low & hi & Medium & mr & Low & sw & Low & tn & Very-Low \\ \midrule
bn & Medium & hu & Medium & mi & Low & sv & Medium & br & Medium \\ \midrule
bs & Low & is & Medium & mn & Low & tg & Low & ns & Very-Low \\ \midrule
bg & Medium & ig & Low & ne & Very-Low & ta & Low & si & Medium \\ \midrule
my & Low & id & Medium & nso & Very-Low & te & Low & yi & Low \\ \midrule
ca & Medium & ga & Low & no & Medium & th & Medium & fy & Medium \\ \midrule
ceb & Low & it & High & ny & Low & tr & Medium & sq & Medium \\ \midrule
zh & Medium & ja & Medium & oc & Very-Low & uk & Medium & ss & Very-Low \\ \midrule
hr & Very-Low & jv & Medium & or & Very-Low & umb & Low & fr & High \\ \midrule
cs & Medium & kea & Very-Low & om & Low & ur & Low & ff & Very-Low \\ \midrule
da & Medium & kam & Very-Low & ps & Low & uz & Very-Low & lo & Low \\ \midrule
nl & Medium & kn & Low & fa & Medium & vi & Medium & lv & Medium \\ \midrule
en & High & kk & Low & pl & Medium & cy & Low & ru & High \\ \midrule
et & Medium & km & Low & pt & High & wo & Very-Low & sr & Medium \\ \midrule
tl & Very-Low & ko & Medium & pa & Low & xh & Low & zu & Low \\ \midrule
fi & Medium & ky & Low & ro & Medium & yo & Low & ba & Low \\ 
\bottomrule
\end{tabular}
\end{adjustbox}
\caption{\label{tab:langtype} ISO-639 code and resource type of languages used in evaluated NMT benchmarks.}
\end{table}

\subsection{FLORES-101}

We use {\tt devtest} subset of FLORES-101 for the evaluation. To better compare different models, we exclude evaluation of language pairs, in which the spBLEU performance of M2M-100 12B~\cite{m2m-100} model is below 3. This gives 5,934 language directions for the comparison. Table~\ref{app:tab:flores-dist} shows the distribution of different categories of language pairs.

\begin{table}[!ht]
\centering
\begin{adjustbox}{width=\linewidth}
\begin{tabular}{lcccccccccccccccc}\toprule
 & VL2VL & VL2L & VL2M & VL2H & L2VL & L2L & L2M & L2H & M2VL & M2L & M2M & M2H & H2VL & H2L & H2M & H2H \\ \midrule
No. lang. pairs & 44 & 200 & 388 & 81 & 144 & 645 & 960 & 186 & 181 & 992 & 1330 & 259 & 35 & 190 & 257 & 42 \\
\bottomrule
\end{tabular}
\end{adjustbox}
\caption{\label{app:tab:flores-dist} Distribution of resource categories for different language directions on FLORES-101~\cite{flores101}.}
\end{table}

\subsection{Tatoeba Challenge}
\label{app:tatoeba:data}

We use the {\tt test} subset data, provided by \citet{tiedemann-2020-tatoeba}\footnote{\url{https://github.com/Helsinki-NLP/Tatoeba-Challenge}} to evaluate all models. We choose a subset of dataset that includes languages which are covered by M2M-100~\cite{m2m-100} model. This brings 1,844 language pairs for the evaluation. The distribution of different language pair categories is shown in Table~\ref{app:tab:tatoeba-dist}. 

\begin{table}[!ht]
\centering
\begin{adjustbox}{width=\linewidth}
\begin{tabular}{lccccccccccccccc}\toprule
 & VL2L & VL2M & VL2H & L2VL & L2L & L2M & L2H & M2VL & M2L & M2M & M2H & H2VL & H2L & H2M & H2H \\ \midrule
No. lang. pairs &  7 & 30 & 37 & 7 & 34 & 144 & 113 & 30 & 144 & 632 & 237 & 37 & 113 & 237 & 42\\
\bottomrule
\end{tabular}
\end{adjustbox}
\caption{\label{app:tab:tatoeba-dist} Distribution of resource categories for different language directions on Tatoeba Challenge.}
\end{table}

\subsection{TICO-19}

We use the evaluation benchmark provided by \citet{anastasopoulos-etal-2020-tico}\footnote{\url{https://tico-19.github.io/}} to compare all models in multilingual medical domain. We utilize language pairs that are included in M2M-100~\cite{m2m-100} model. This gives us 650 language pairs for the evaluation. Table~\ref{app:tab:tico19-dist} shows the number of language pairs in different resource types of language directions.

\begin{table}[!ht]
\centering
\begin{adjustbox}{width=\linewidth}
\begin{tabular}{lcccccccccccccccc}\toprule
 & VL2VL & VL2L & VL2M & VL2H & L2VL & L2L & L2M & L2H & M2VL & M2L & M2M & M2H & H2VL & H2L & H2M & H2H \\ \midrule
No. lang. pairs &  12 & 48 & 24 & 16 & 48 & 132 & 72 & 48 & 24 & 72 & 30 & 24 & 16 & 48 & 24 & 12 \\
\bottomrule
\end{tabular}
\end{adjustbox}
\caption{\label{app:tab:tico19-dist} Number of language pairs for different categories of language directions on TICO-19.}
\end{table}

\section{Hyper-Parameters for Architecture and Optimization}
\label{app:impl}

\begin{table}[!ht]
\centering
\begin{adjustbox}{width=0.8\linewidth}
\begin{tabular}{lc|lc}\toprule
Hyper-Parameter & Specification & Hyper-Parameter & Specification\\ \midrule
Encoder Layer & 12 & Scheduler & inverse-sqrt\\
Decoder Layer & 3 & Optimizer & Adam \\
Encoder Emb dim. & 1024 & Clip Norm & 1.0 \\
Decoder Emb dim & 1024 & Learning Rate & 1e-4 \\
Encoder FFN Emb dim. & 4096 & Warmup init LR & 1e-07 \\
Decoder FFN Emb dim. & 4096 & Warmup Updates & 40K \\
Number of attn heads & 16 & Adam Betas & 0.9,0.98\\
Attention Dropout & 0.1 & Adam eps & 1e-6 \\
Share Encoder/Decoder Emb. & True & Label Smoothing & 0.1 \\
FP16 & True & Dropout & 0.1 \\
Loss Scalar & 2.0 & Max Tokens (per GPU) & 1,000 \\

\bottomrule
\end{tabular}
\end{adjustbox}
\caption{\label{app:tab:hyper-params} List of hyper-parameters used for the architecture, and optimization. }
\end{table}

\newpage 
\section{spBLEU Results}
\label{app:spbleu}
~~
\subsection{spBLEU of Remaining Language Directions}
\label{app:spbleu:remain}

\begin{table}[!ht]
\centering
	\begin{adjustbox}{width=0.6\textwidth}
	\centering
	\begin{tabular}{lcccccc}\toprule
		 &\multicolumn{5}{c}{\hspace{3.5cm} Language Direction} \\
		\cmidrule{4-7} 
		Model & params & Speed & M2M & M2H & H2M & H2H \\ \midrule
		\textbf{FLORES-101} \\[1ex] 
		FLORES-124 & 175M & 5.3$\times$ & 13.7 & 18.0 & 16.8 & 23.0 \\
		M2M-100 & 418M & 3.1$\times$ & 18.1 & 23.0 & 21.6 & 27.8 \\
		FLORES-124 & 615M & 2.9$\times$ & 19.3 & 24.1 & 22.7 & 28.9 \\
	    \textit{Finetuned-100} & 330M & 7.8$\times$ & 16.4 & 21.2 & 19.7 & 26.0 \\
	    \textit{SMaLL-100} & 330M & 7.8$\times$ & 19.3 & 24.2 & 22.6 & 28.8 \\
	    
	    \midrule
	    M2M-100 & 1.2B & 1.8$\times$ & 22.3 & 28 & 25.8 & 32.7 \\
	    M2M-100 & 12B & 1$\times$ & 23.9 & 29.5 & 27.6 & 34.3 \\		\midrule \midrule

	   \textbf{Tatoeba} \\[1ex] 
		FLORES-124 & 175M  & 5.3$\times$ & 25.2 & 28.1 & 24.5 & 35.3 \\
		M2M-100 & 418M  & 3.1$\times$ & 29.6 & 34.0 & 29.6 & 41.6 \\
		FLORES-124 & 615M  & 2.9$\times$ & 31.7 & 35.5 & 31.0 & 43.0 \\
	    \textit{Finetuned-100} & 330M  & 7.8$\times$ & 28.3 & 34.2 & 28.1 & 39.2 \\
	    \textit{SMaLL-100} & 330M  & 7.8$\times$ & 31.9 & 36.3 & 31.1 & 42.4 \\
	    \midrule
	    M2M-100 & 1.2B  & 1.8$\times$ & 33.8 & 39.0 & 34.2 & 47.4 \\
	    M2M-100 & 12B & 1$\times$ & 33.1 & 39.0 & 34.2 & 48.7 \\ \midrule \midrule

	    \textbf{TICO19}  \\[1ex] 
		FLORES-124 & 175M & 5.3$\times$ & 15.1 & 20.3 & 17.7 & 28.1 \\
		M2M-100 & 418M & 3.1$\times$ & 20.6 & 26.6 & 23.6 & 33.1 \\
		FLORES-124 & 615M & 2.9$\times$ &  20.2 & 27.0 & 23.3 & 33.9 \\
	    \textit{Finetuned-100} & 330M & 7.8$\times$ & 19.7 & 24.4 & 23.4 & 31 \\
	    \textit{SMaLL-100} & 330M & 7.8$\times$& 21.7 & 27.4 & 25.2 & 33.7 \\ 
	    \midrule
	    M2M-100 & 1.2B & 1.8$\times$ & 21.1 & 30.2 & 24.8 & 37.7 \\
	    M2M-100 & 12B & 1$\times$ & 24.8 & 33.1 & 28.3 & 39.4 \\
	    \midrule
		\bottomrule
	\end{tabular}
	\end{adjustbox} 
	\caption{ Average spBLEU performance of different models on FLORES-101, Tatoeba, and TICO-19 benchmarks for medium and high-resource language pair categories, defined in Section~\ref{sec:result}. The FLORES-101 results are computed on language pairs where M2M-100 12B has spBLEU scores higher than 3 to avoid polluting the analysis with meaningless scores. The first and second columns give the model size and speed-up ratios compared to M2M-100~(12B).}
\end{table}

\newpage
\subsection{FLORES-101}

﻿\begin{table}[!ht]
    \centering
    \begin{adjustbox}{width=\linewidth}
\begin{tabular}{|l|l|l|l|l|l|l|l|l|l|l|l|l|l|l|l|l|l|l|l|l|l|l|l|l|l|l|l|l|l|l|l|l|l|l|l|l|l|l|l|l|l|l|l|l|l|l|l|l|l|l|l|l|l|l|l|l|l|l|l|l|l|l|l|l|l|l|l|l|l|l|l|l|l|l|l|l|l|l|l|l|l|l|l|l|l|l|l|l|l|l|l|l|l|l|l|l|l|l|l|l|l|}
    \hline
        \diagbox{src}{tgt}  & afr & amh & ara & hye & asm & ast & azj & bel & ben & bos & bul & mya & cat & ceb & zho & hrv & ces & dan & nld & eng & est & tgl & fin & fra & ful & glg & lug & kat & deu & ell & guj & hau & heb & hin & hun & isl & ibo & ind & gle & ita & jpn & jav & kea & kam & kan & kaz & khm & kor & kir & lao & lav & lin & lit & luo & ltz & mkd & msa & mal & mlt & mri & mar & mon & npi & nso & nob & nya & oci & ory & orm & pus & fas & pol & por & pan & ron & rus & srp & sna & snd & slk & slv & som & ckb & spa & swh & swe & tgk & tam & tel & tha & tur & ukr & umb & urd & uzb & vie & cym & wol & xho & yor & zul \\ \hline
afr & -1 & 8.9 & 20.7 & 10.6 & -1 & 22.2 & 7.2 & 10.9 & 22.4 & 24.2 & 30.8 & 2.1 & 32.8 & 14.1 & 15.7 & 23.4 & 25.5 & 34.3 & 24.5 & 49.2 & 23.2 & 20.6 & 19.1 & 34 & -1 & 27.1 & -1 & 7.2 & 27.9 & 22.8 & 4 & 9 & 22.6 & 26.1 & 22.4 & 16 & 10.9 & 31.2 & -1 & 23.9 & 18.9 & 17.8 & -1 & -1 & 1.8 & 7.8 & 8.7 & 15 & -1 & 7.9 & 11.1 & -1 & 22.9 & -1 & 18.4 & 28.8 & 28.7 & 15 & -1 & -1 & 11.3 & 2.9 & -1 & -1 & 27 & -1 & 24.4 & -1 & -1 & 6.3 & 21.6 & 17.9 & 34.4 & 7.4 & 29.9 & 22.7 & 26.4 & -1 & 6.1 & 26.6 & 24.1 & 1.2 & -1 & 22.5 & 26.2 & 32 & -1 & 7.4 & -1 & 4.5 & 23.2 & 23.4 & -1 & 13.7 & -1 & 30.3 & 6 & -1 & 3.8 & -1 & 6.5 \\
\hline
    \end{tabular}
\end{adjustbox}
\caption{spBLEU performance of last checkpoint of SMaLL-100 model on language pairs of FLORES-101.}
\end{table}

\subsection{Tatoeba}

﻿\begin{table}[!ht]
    \centering
    \begin{adjustbox}{width=\linewidth}
    \begin{tabular}{|l|l|l|l|l|l|l|l|l|l|l|l|l|l|l|l|l|l|l|l|l|l|l|l|l|l|l|l|l|l|l|l|l|l|l|l|l|l|l|l|l|l|l|l|l|l|l|l|l|l|l|l|l|l|l|l|l|l|l|l|l|l|l|l|l|l|l|l|l|l|l|l|l|l|l|l|l|l|l|l|l|l|l|l|l|l|l|l|l|l|l|l|l|l|l|l|l|l|l|l|l|}
    \hline
        \diagbox{src}{tgt} & af & am & ar & ast & ba & be & bg & bn & br & ca & ceb & cs & cy & da & de & el & en & es & et & fa & ff & fi & fr & fy & ga & gd & gl & gu & ha & he & hi & ht & hu & hy & ig & ilo & is & it & ja & jv & ka & kk & km & kn & ko & lb & lg & ln & lo & lt & lv & mg & mk & ml & mn & mr & ms & my & nl & oc & pa & pl & ps & pt & ro & ru & sd & si & sl & so & sq & su & sv & ta & th & tl & tn & tr & uk & ur & vi & wo & xh & yi & yo & zh & zu \\ \hline
af & -1 & -1 & 6.9 & -1 & -1 & -1 & -1 & -1 & -1 & -1 & -1 & 12 & -1 & 78.3 & 41.9 & -1 & 47.3 & 48.9 & 19.9 & -1 & -1 & -1 & 47.8 & 12.4 & -1 & -1 & -1 & -1 & -1 & 23.6 & -1 & -1 & -1 & -1 & -1 & -1 & 48.1 & 54.3 & 39.7 & -1 & -1 & -1 & -1 & -1 & -1 & 5.2 & -1 & -1 & -1 & -1 & -1 & -1 & 16.7 & -1 & -1 & -1 & 16.1 & -1 & 51.3 & -1 & -1 & 60.5 & -1 & 43.2 & 41.7 & 34.3 & -1 & -1 & -1 & -1 & -1 & -1 & 89.1 & -1 & -1 & -1 & 1.8 & 36.6 & 33.4 & -1 & 85.8 & -1 & 1.8 & 4.9 & -1 & 21.5 & 4 \\ \hline
am & -1 & -1 & -1 & -1 & -1 & -1 & -1 & -1 & -1 & -1 & -1 & -1 & -1 & -1 & -1 & -1 & 35.1 & -1 & -1 & -1 & -1 & -1 & -1 & -1 & -1 & -1 & -1 & -1 & -1 & -1 & -1 & -1 & -1 & -1 & -1 & -1 & -1 & -1 & -1 & -1 & -1 & -1 & -1 & -1 & -1 & -1 & -1 & -1 & -1 & -1 & -1 & -1 & -1 & -1 & -1 & -1 & -1 & -1 & -1 & -1 & -1 & -1 & -1 & -1 & -1 & -1 & -1 & -1 & -1 & -1 & -1 & -1 & -1 & -1 & -1 & -1 & -1 & -1 & -1 & -1 & -1 & -1 & -1 & -1 & -1 & -1 & -1 \\
\hline
    \end{tabular}
\end{adjustbox}
\caption{spBLEU performance of last checkpoint of SMaLL-100 model on language pairs of Tatoeba.}
\end{table}

\newpage
\subsection{TICO19}

﻿\begin{table}[!ht]
    \centering
    \begin{adjustbox}{width=\linewidth}
    \begin{tabular}{|l|l|l|l|l|l|l|l|l|l|l|l|l|l|l|l|l|l|l|l|l|l|l|l|l|l|l|}
    \hline
        \diagbox{src}{tgt} & am & ar & bn & my & zh & en & tl & fr & lg & ha & hi & id & km & ln & ms & mr & ne & ps & fa & ru & so & es & sw & ta & ur & zu\\ \hline
am & -1 & 7.5 & 12.7 & 1.5 & 7.6 & 18.3 & 11.3 & 13.7 & 2.4 & 6.9 & 16.2 & 14.2 & 5.5 & 2.4 & 14.4 & 5.9 & 6.1 & 7.3 & 11.8 & 10.6 & 0.5 & 15.7 & 13.7 & 6 & 10.4 & 5.2 \\ \hline
ar & 7.8 & -1 & 16.8 & 1.5 & 12.3 & 29.9 & 3.8 & 23.6 & 1.5 & 7.4 & 24.9 & 26 & 9.8 & 1.8 & 23.9 & 4.2 & 0.3 & 6.4 & 20.5 & 18.8 & 0.4 & 29.5 & 21.3 & 1.7 & 13.3 & 2.5 \\ \hline
bn & 9.7 & 15.5 & -1 & 4.4 & 13.8 & 33.8 & 23.9 & 21.6 & 3.9 & 10.7 & 31.1 & 27.9 & 12.6 & 4.4 & 27.3 & 11.3 & 16.4 & 10.4 & 21.1 & 19.4 & 1.6 & 27.9 & 22.7 & 9.9 & 18.3 & 9.9 \\ \hline
my & 6.7 & 6.3 & 11.4 & -1 & 7.9 & 16.6 & 14.2 & 11.2 & 3.2 & 7.9 & 14.7 & 14.3 & 7.9 & 2.8 & 14.8 & 5.6 & 8 & 7.9 & 11.5 & 10.2 & 1.3 & 14.4 & 13.7 & 6.1 & 11 & 7.1 \\ \hline
zh & 8.6 & 15.6 & 18.3 & 4 & -1 & 27.5 & 20.6 & 20.9 & 3.6 & 9.3 & 24.3 & 26.5 & 10.6 & 4 & 24.5 & 8.5 & 6.8 & 9.2 & 19.9 & 19 & 1.4 & 26.6 & 20.9 & 5.3 & 15.2 & 9 \\ \hline
en & 11.1 & 25.2 & 26.8 & 7 & 19.6 & -1 & 35.6 & 36.3 & 5.6 & 15.4 & 39.3 & 47 & 17.9 & 3.6 & 44.7 & 12.1 & 19 & 11.8 & 29.9 & 29.7 & 2.1 & 47.5 & 32.4 & 9.3 & 20.8 & 15.7 \\ \hline
tl & 10.1 & 15.2 & 21.4 & 6 & 14.6 & 45.4 & -1 & 26.1 & 5.7 & 14 & 28.5 & 33.9 & 14.7 & 6 & 33.2 & 10.1 & 14.6 & 10.6 & 21.9 & 22.5 & 1.6 & 33.3 & 26.2 & 7 & 17 & 14.6 \\
\hline

    \end{tabular}

\end{adjustbox}
\caption{spBLEU performance of last checkpoint of SMaLL-100 model on language pairs of TICO19.}
\end{table}

\section{Details of Fine-Tuning on Low-Resource Language Pairs}
\label{app:detail-low}

For further fine-tuning of SMaLL-100 and M2M-100~(418M) models on selected language pairs, we use bilingual data provided by \citet{tiedemann-2020-tatoeba}~(release 2021.08.07)\footnote{\url{https://github.com/Helsinki-NLP/Tatoeba-Challenge/blob/master/data/README-v2021-08-07.md}} as its training data is less noisy. We evaluate fine-tuned models on {\tt devtest} subset of FLORES-101~\cite{flores101} benchmark with spBLEU metric~\cite{flores101}. We use the same hyper-parameters, as defined in Appendix~\ref{app:impl}. We train each model on 2 TESLA V100-32GB GPUs.

\end{appendices}

\end{document}